# Visualization of Tradeoff in Evaluation

## *from Precision-Recall & PN to LIFT, ROC & BIRD*

**David M.W. Powers,** Beijing University of Technology, China & Flinders University, Australia
*Technical Report KIT-14-002 Computer Science, Engineering & Mathematics, Flinders University*

***Abstract*** – Evaluation often aims to reduce the correctness or error characteristics of a system down to a single number, but that always involves trade-offs. Another way of dealing with this is to quote two numbers, such as Recall and Precision, or Sensitivity and Specificity. But it can also be useful to see more than this, and a graphical approach can explore sensitivity to cost, prevalence, bias, noise, parameters and hyper-parameters.

Moreover, most techniques are implicitly based on two balanced classes, and our ability to visualize graphically is intrinsically two dimensional, but we often want to visualize in a multiclass context. We review the dichotomous approaches relating to Precision, Recall, and ROC as well as the related LIFT chart, exploring how they handle unbalanced and multiclass data, and deriving new probabilistic and information theoretic variants of LIFT that help deal with the issues associated with the handling of multiple and unbalanced classes.

***Keywords*** – Multi-class evaluation, Multiclass visualization, Chance-correct evaluation, Receiver Operating Characteristics, Area Under the Curve, Recall, Precision, Accuracy, Kappa

## Introduction

How do we decide how good a system is?

There are basically two approaches – we can see how often it is correct or we can look at how often it is wrong, but it doesn't really matter which way we go on this. If we ignore the classes and just consider the marks, right or wrong, we have a simple accuracy or error rate – usually expressed as a probability or a percent correct, just like marks for an exam. While classifiers often work on minimizing error, we tend think in terms of how good we are rather than how bad. We will therefore focus on goodness measures, but for every one of these there will be a complementary error measure.

This approach is simplistic in several ways. In particular, some problems may be harder than others, some classes may be not as easy to predict as others, and indeed some situations may be very rare or very significant or very costly, and deserve more weight than others. The problem of evaluation in machine learning has become topical recently, with a series of AAAI and ICML workshops [1,2], as well as considerable interest in graphical or visualization methods to assist in more intuitive understanding of the tradeoffs and avoid relying on a single number [3]. However these techniques still tend to make the assumption that we are dealing with a two-class problem, and even work that does seek to extend ROC to the multi-class problem tends to have the aim of reducing the graphical representation to a single number, as a simple average of Area Under the Curve (AUC) across a set of dichotomous classifiers [4,5,6] or a the Volume Under the (hyper)Surface (VUS) of a multidimensional generalization of AUC [7,8].

Area Under the Curve has a meaningful interpretation in terms of the probability of correctly ranking a pair of instances [9], but AUC and variants spread weight over all possible operating points of a classifier, even though to deploy it in a specific case for classification requires choosing a single operating point. AUC has been characterized as having two components: a certainty component related to classification at an optimal operating point, and consistency component related to fall off as conditions change [10].

In this paper we will review the case where there are only one or two classes of interest, like the class of relevant documents, or a set of yes/no questions. We will then consider the impact of the prevalence and cost of particular classes, and the complementary question of bias in the classifier. In addition we consider the desirability of predicting the right number of instances for each label, the impact of imbalances and errors/noise in the data, and their implicit relationship with cost.

We will then concentrate on the PR, ROC and LIFT visualizations, and review their various advantages and disadvantages, including those of the derivative measures that have been proposed to improve them. Finally, we will move to the multiclass case which is our primary focus, where we will see that there are some very difficult issues, and we will present our proposed modifications to address the problems, and produce clear and meaningful multiclass visualizations, rather than just a new single point of failure as arises when all of this information is reduced to a single number, irrespective of which relative of Accuracy, Kappa, AUC or VUS it takes.

# I. The Dichotomous Case

The basic measures used for evaluation are usually based on a dichotomous distinction, right or wrong, relevant or irrelevant, positive or negative. We assume here the supervised learning paradigm distinguishing positives and negatives based on a dataset containing examples of both, or the corresponding use of such a dataset for evaluating a hand-designed or hand-tuned system. Strictly speaking there should always be at least two classes being classified, but sometimes we have a focus on a single class in terms of our evaluation (e.g. use of Recall, Precision and F-measure in Information Retrieval [11]) or learning (e.g. novelty detection with one-class SVM [12]). Rand Accuracy on the other hand takes into account accuracy for both classes and is definable directly from a contingency table (Table 1) or in terms of weighted averages of the single class accuracy Precision (this is about how accuracy predictions are) or the single class Recall (this is about how successful we are at discovering class members).

Recall (R or Rec) and Precision (P or Prec) are defined in terms of one class termed positive (+ve) in terms of true positives (tp=TP/*N*), positive prevalence (rp=Prev) and negative bias (pp=Bias). Here prevalence is the relative frequency or empirical probability of a class, and bias is the relative frequency or sample probability of a prediction. Table 1 explains a standard systematic notation based on a contingency table showing predicted labels horizontally (**±P**) and real labels vertically (**±R**), where the left hand table shows probabilities. Thus on the left, the four coloured cells sum to 1, as do the horizontal and vertical margins which correspond respectively to the prevalences of and biases towards the corresponding class. The right hand table shows raw counts that are dependent on sample size so the four coloured cells sum to *N,* the size of the test set, as do the horizontal and vertical margins. Recall and Precision also have alternate names in this systematic notation: tpr for true positive rate and tpa for true positive accuracy (sometimes upper case is used).

**Table 1. Systematic notation used *sans serif* in our formal equations (1-10). [1,3]**

| | **+R** | **−R** | |
|---|---|---|---|
| **+P** | tp | fp | pp |
| **−P** | fn | tn | pn |
| | rp | rn | 1 |

| | **+R** | **−R** | |
|---|---|---|---|
| **+P** | TP | FP | PP |
| **−P** | FN | TN | PN |
| | RP | RN | *N* |

For completeness we present the equations for Recall (Rec) and Precision (Prec) and define F-measure (1-3) using either the harmonic mean (*hm*) of Recall and Precision, or equivalently expressing our true positives relative to an assumed common distribution represented by the arithmetical mean (*am*) of rp (Prev) and pp (Bias). We also show (4) the Rand Accuracy (Acc) as it is calculated from the two forms of the contingency table. However, it is easy to see that Accuracy is a prevalence-weighted average of Recall (of positives) and Inverse Recall (which relates similarly to success in finding negatives), and that conversely, it is also a bias-weighted average of Precision (of positive predictions) and Inverse Precision (precision of negative predictions). Formally:

$$\text{Rec} = \mathsf{TP} / \mathsf{RP} = \mathsf{tp} / \mathsf{rp} = \mathsf{tpr} = \text{Se} \qquad (1)$$

$$\text{Prec} = \mathsf{TP} / \mathsf{PP} = \mathsf{tp} / \mathsf{pp} = \mathsf{tpa} \qquad (2)$$

$$\text{F1} = \mathsf{tp}/am(\mathsf{rp},\mathsf{pp}) = hm(\text{R},\text{P}) \qquad (3)$$

$$\text{Acc} = [\mathsf{TP}+\mathsf{TN}]/N = \mathsf{tp}+\mathsf{tn} \qquad (4)$$

Recall is such an important measure that there are whole families of other names for it and its inverse and complementary forms, and in some fields it is better known as Sensitivity (Se). In addition, the most important graphical tradeoff methods are based on the Recall and family, including ROC, LIFT and Precision-Recall (PR) graphs. However PR graphs are always biased, although they are often advocated as providing clearer detail in highly skewed applications like Information Retrieval, with $\text{skew}_{\text{Prev}} = \mathsf{RN}/\mathsf{RP} \gg 1$ and $\text{skew}_{\text{Bias}} = \mathsf{PN}/\mathsf{PP} \gg 1$ (relatively few are relevant, relatively few are predicted). In this case $0 \lesssim \mathsf{rp} \simeq \mathsf{pp} \simeq \mathsf{tp} \ll 1$ and $0 \ll \text{tnr} \lesssim 1$. That is, Inverse Recall is close to 1, and Recall and Precision are concerned with very tiny proportions of the instances.

In the systematic naming system of Table 1, we noted that tpr is the true positive rate, the rate at which true positives are correctly predicted positive, and similarly tnr is the true negative rate, the rate at which true negatives are correctly predicted negative. These *rates* can be interpreted as *proportions* of positives and negatives correctly identified, or equivalently the *probability* that positives resp. negatives are correctly identified. The *rates* relate to the *real* labels (how many of them are *found*), while accuracy and precision evaluate with respect to the *predicted* labels (how many of them are *correct*).

There are other rates relating to errors, including in particular the false positive rate fpr, the rate at which true negatives turn up falsely as positive predictions, and the false negative rate, fnr, the rate at which true positives turn up as negatives.

Recall = tpr = Sensitivity (1) and for the negatives we analogously define Inverse Recall = tnr (5), which is also known as Specificity (Sp), and can be ignored in Information Retrieval because it is close to 1 due to the number of predictions being so much smaller than the number of irrelevant documents, so there is not much scope for optimization. This is not the case for Intelligent Systems in general. Fallout = fpr = 1–Sp (6) is a way of characterizing the contamination of our positive predictions by a proportion of the negative examples:

InvRec = TN / RN = tn / rn = tnr = Sp (5)

Fallout = FP / RN = fp / rn = fpr = 1-Sp (6)

Sensitivity can be plotted against Specificity, but it is more common to plot a reflection of this: tpr vs fpr. This is the well-known Receiver Operating Characteristics (ROC). Closely related is the PN (Positive-Negative) chart which directly plots TP against FP without normalizing to probabilities by dividing by the number of positives or negatives. Such linear scalings won't change the appearance of an autoscaled graph…

Given we don't constrain the axes to the same scale, the PN graphs look identical and we need only change the values we label the axes with. However one caution is that the angles of lines drawn on the graph change if axes are not drawn with equal scale. In particular the 45° positive diagonal joining (0,0) and (1,1) in ROC is of special significance (chance-line) [1,3] but the 45° negative diagonal joining (RP,0) and (0,RP) in PN is also of interest (it is known in Information Retrieval as the break-even line).

### *PR & PN – the Break-Even Case*

The break-even case arises when Bias = Prevalence viz. RP = PP and hence RN = PN and FN = FP and Recall = Precision. It means we have made the right number of predictions for each class, whereas a poorly biased system will tend to overpredict one class and underpredict another. This is thus a useful heuristic when the false positives and false negatives are regarded as of equal weight or cost and encompass similar variance or noise. Clearly for a perfect score the number of real positives and predicted positives must match and some but not all classifiers tend to achieve this tradeoff. The Recall =

Precision line on a PR graph, and the FP + TP = 1 line on a PN graph, represent this tradeoff point, where the curve crosses the line. This is illustrated for the PR and PN graphs of Fig.1 (where the average skew is of course 6 across the six classes plotted, so the PN x-intercept for the break-even lines varies around 1/6 or 16.7% on the ROC-scale according to the actual prevalences shown in the key).

### *ROC – the Chance Case*

The chance case arises when no informed prediction is made but we have pure guessing. In this case we expect to see positive predictions turning up at the same rate for both positive instances (tpr) and negative instances (fpr). The chance line is thus the tpr=fpr positive diagonal through the origin in the ROC curve, and all isobars parallel to the chance line represent classifiers with equal levels of informedness or guessing.

Well known in statistics as Youden's J, Informedness J = κɪ = ΔP' = ΔR = B is tpr–fpr = Rec + InvRec – 1 and represents the distance above the chance line, being formally the probability of an *informed* prediction (as opposed to a guess) for a specific system. In the context of ROC analysis, J was independently shown [13] to be a skew independent measure of weighted relative accuracy (WRAcc), and its average across all parameterizations represented by the ROC curve is Gini, while the total area under the ROC curve (AUC) is closely related to the Mann-Whitney U statistic, which is the corresponding area under the PN curve. In Psychology the same formula emerged as the pair of regression coefficients ΔP and ΔP' which model well the directional strengths of association measured in human subjects, while their geometric mean corresponds to Matthews Correlation [14].

Maximizing the distance of an operating point above the chance curve, is equivalent to maximizing its perpendicular distance above and left of the chance line, and thus finding the nearest tangential isocost line of the perfect (0,1) point, and maximizes Informedness. ROC and Informedness can be related to cost under the assumption that the positives and negatives have equal cost in total (balanced iff fnr=fpr), while break-even assumes that positives and negatives each have equal cost (balanced iff FN=FP).

The same Informedness quantity is also known in gambling and trading as an *edge*, and given standard bookmaker odds the expected winnings given by the Bookmaker (B) cost formula correspond to the multiclass informedness. Here the value of a bet (the cost to punter of a loss or bookmaker of a win) is measured in terms of the odds of winning, that is balancing the expected number of wins and losses, and giving them equal cost in total. Informedness is also known in education, where the same multiclass formula is used to ensure fair marking of multiple choice exams, independent of the number of choices: the odds of guessing wrong for a set of questions with K possibilities each are K-1:1 and so weights for right and wrong answers are set accordingly (e.g. 3 marks for a right answer, -1 for a wrong answer, for K=4). It is also sometimes useful to look at the information flow in the other direction, or Markedness = ΔP = Prec + InvPrec – 1, that is how often the predicting variable is actually *marked* by the real class/situation. As noted earlier, the geometric mean of Informedness and Markedness is Matthews Correlation. We also discuss below application of the Bookmaker accounting in other costing frameworks, in particular an information theoretic costing giving a formula for expected information gain, each costing in general leading to a different ideal operating point.

In the break-even case, we not only have Recall = Precision = F-measure but Informedness = Markedness = Correlation = Kappa [15] (representing a variety of other chance-corrected measures including the well known Fleiss Kappa and Cohen Kappa). Informedness and Markedness are indeed shown [15] to also be representable as Kappa-renormalizations of Recall and Precision after subtracting off the component due to chance. The Area under Kappa (AUK) has been proposed as an alternative to ROC AUC, originally using Cohen Kappa [16]. Cohen and Fleiss Kappa [15] do not have known probabilistic interpretations, and deviate strongly from Informedness as bias varies from prevalence. Like Bookmaker Informedness they are also defined for multiple classes rather than just the dichotomous case considered so far. The only consistent Kappa is $\kappa_B = \kappa_I$ = Bookmaker Informedness [15], and this form of AUK is seen in Fig. 1 as the area under the Bookmaker Operating Characteristics curve (BOC), and equates to Gini/2.

Note that the occurrence of errors means that the full complement of positive cases is reached after more than the optimal number of positive predictions, but the same can be said for negative predictions (the ROC curve achieved by flipping both axes). But when the data becomes unbalanced the same fpr (resp. fnr) will insert more errors into the smaller class and less into the bigger class, and shift the operating point right or left, leading to an increase or decrease in the setting of the bias towards predicting the class. Similarly for noise imbalance, e.g. a label noise model that introduces more error for one class, will lead to slower discovery of that class and an increase in the operating bias. A further case where this happens is the multiclass case. When there are more than two classes, there are always going to be imbalances for one (+ve) vs rest (–ve) evaluation.

### *AUK – the Balanced Case*

We now look at how well common evaluation measures are interpretable across different classifiers and datasets, including the same diagnosis or prediction under different demographic or environmental conditions. For the chance-correct or chance-corrected measures, they are by definition expected to be 0 in the face of pure guesswork, or predictions that are independent of (or uncorrelated with) the actual class. This condition does not hold for uncorrected measures (Accuracy, Recall, Precision or F1).

If we consider our K-way multiple choice exam, the probability of a student being correct is 1/K and this is the expected accuracy by chance, while the other measures consider the accuracy of only a single class and can be biased very high, and if we are only trying to predict whether a student will guess a question right or wrong we can achieve an accuracy of 1-1/K by predicting that they will always get it wrong. Of course, the accuracy of this prediction will be lower as the students are more informed.

There are two ways of dealing with this problem:

a. ensure that our training sets are binary and balanced –the only case in which these standard uncorrected measures are unbiased, as the chance level is 0.5 for both classes;

b. subtract off the expected score due to chance, and renormalize to the form of a probability by dividing by the expected error, the room for improvement: κ=[Acc-ExpAcc]/ExpErr.

The Kappa view of Informedness [15] is κI = [Recall-Bias]/[1-Prevalence], which for our 'always wrong' prediction of outcomes in a K-way exam will correctly give us 0. On the other hand, the variant of Kappa based on our naïve expectation of 1/K chance for the always-wrong guessing is [Accuracy-ExpAcc]/[1-ExpAcc] = [1-2/K]/[ 1-1/K]. This is [K-2]/[K-1] $\backsimeq$ 1-1/K for K >> 1. What distinguishes the different Kappa measures is the model of expectation, and Informedness (ΔP) is arguably appropriate in the absence of costings contradicting the default of balancing the cost of the negatives against the cost of the positives [3,4,5]. We will consider the case where we take account of other cost ratios.

However, in the balanced prevalence case, we no longer have any difference between balancing the cost of individual positives and negatives and balancing the cost of all positives and all negatives, or any difference between the marginal distributions. Thus in this case optimizing Accuracy and optimizing any Kappa gives the same result as optimizing Informedness. Nonetheless, a specific cost allocation can lead to a different optimum even when balanced, and conversely misoptimization may occur when the balanced data set does not reflect the natural prevalences for the environment where the system will be deployed.

### *BOC Curves*

Plotting Kappa, and considering the area under the Kappa curve (AUK), has been recommended as more meaningful than ROC and ROC AUC [16]. Naturally, we can plot any of the Kappa variants, and use it for calculating AUK. This is illustrated as BOC (Bookmaker Operating Characteristics) in Fig. 1 for $\kappa_I$, but different choices give different curves that coincide only for this balanced case, and give rise to different values for AUK due to the different bias-prevalence tradeoffs as they move out of balance, with the AUK for Bookmaker Informedness (BOC) being well known as Gini [13], and weighting the loss and cost relative to the probabilities or winning and losing (according to the standard approach to bookmaker odds) [15].

### *ROCH Curves*

In real life, one type of error may cost more than another. In many applications there are missed positive (false negative) errors that could lead to loss of life or massive costs, while the false alarm (false positive) errors may give us a scare and add a small cost. The assumption that a false positive costs the same as a false negative is implicit in the common evaluation measures (1-4) as well as the complementary error or distance measures (e.g. Err = 1–Acc). Equating the cost of getting all the positives wrong with that of getting all the negatives wrong is done by design in ROC or Informedness, with individual costs that are inversely proportional to prevalence, viz. $cp \propto 1/rp$ and $cn \propto 1/rn$.

Thus the ROC chance line with gradient given by $tpr/fpr = [tp/rp]/[fp/rn] = 1$, is a zero-cost line in the group costing sense. In a PN graph this is $TP/FP = RP/RN$. These lines through the origin can be rewritten in terms of the individual costs attributed as $tp/fp = cn/cp$ or $tp*cp + tn*cn = cn$ (e.g. at the origin we have 0% true positives and 100% true negatives with their associated cost). Changing the costs $cp$ and $cn$ from the ROC-implied defaults changes the gradient of the equal cost lines. These are also known as isocost contours in general, and are straight parallel lines in ROC and PN charts, but not for other charts such as Precision-Recall (PR) charts [13].

Thus ROC charts can be easily be used to explore changes in costs or equivalently changes in prevalences. The sharper the elbow at the optimum, the bigger the change in cost gradient ($skew_{Cost}$) that is needed to shift it to a different operating point.

ROC charts surrounding the area under the curve (AUC) with a convex hull (rubber band) are called ROCH or ROCCH charts and basically connect pareto-optimal points with a straight line segment. These points are optimal in some sense or under some assumed skew, and in Computational Intelligence points on the Pareto front (frontier) are said to dominate points that are below the front (the hull in ROCCH terminology). Clearly all the points on such a segment are equally good for the cost or prevalence skew corresponding to its gradient. Conversely points that are not on the hull, but inside the surrounded area, can't be an optimum for any cost or prevalence variant. In addition, any point on a segment is an achievable interpolation of the systems that correspond to its endpoints – for example one can randomly choose which system to use with a probability inverse to their distance from the target point.

This fusion of systems (hull) allows choosing between points of equivalent cost based on other considerations, such as balancing the number of false positives and false negatives or their rates, but in cost or informedness terms all systems defined by the segment are equivalent.

There is however a caution that given the ROC or ROCH chart is displaying test data, and the fusion is based on that test data, there is *no* guarantee that the interpolating points on the hull will be better for independent data – the zigs and zags may be noise.

This understanding of the PN, ROC and ROCH curves in terms of prevalence and cost allows us a deeper understanding of the areas defined by them, in particular ROCH AUC (and related measures like Gini). These averaged measures have two components, how good they are at the operating point selected, based on current costs and prevalences, and how robust they are to changes in these costs and prevalences. We can regard these two components as representing Certainty (or for Gini, Informedness) and Consistency (which is represented by the area between the curve and the triangle defined by the selected operating point and the chance line) [10].

## II. The Distributional Variants

One of the main controversies about trade-off graphs is the appropriateness of the distribution implied by the choice of variable on the x-axis, with Recall=tpr=Sensitivity being assumed on the y-axis for PR, ROC and LIFT charts. We have already looked at the AUK variant of ROC AUC. Whereas ROC plots a chance corrected measure rather than Recall on the y-axis, against Fallout=fpr=1-Specificity, AUK plots Kappa, and the most directly interpretable, which we advocated above, is $\kappa_I$ = B = tpr–fpr = ΔR.

### *The H-curve*

Hand [17] notes that the ROC AUC is averaging tpr over a distribution that is dependent on the classifier rather than just distribution of classes in the real world. He considers that a uniform distribution doesn't reflect the natural bias towards a particular prevalence or cost for each class, and proposes a Beta distribution as a model of the cost associated with each class. In its most general form, like the F-measure, it provides the opportunity to weight the positive and negative components, but rather than being based on a harmonic mean (3) it is related to the geometric mean ($gm(x,y)=(x*y)^{1/2}$). We consider here only the equally weighted form that is effectively averaged (integrated) over cp to define the H-measure (which in the original paper refers not to the plotted function but to the area under the modified curve, which we denote by H-AUC).

Given the equal cost of *all* positives and *all* negatives case associated with ROC [13], note that we have an inverse relationship with prevalence: $cp \propto 1/rp$ and $cn \propto 1/rn$, giving the equivalence (apart from constant factors) that we show in (7). Interestingly all the different chance correct measures analysed (Informedness, Markedness, Correlation, Fleiss and Cohen Kappas) can be shown to be just different normalizations of the determinant |C| by different ways of averaging prevalences and/or biases [15]. Thus

$$H \propto tpr * gm_2(cp,cn) \propto tpr / gm_2(rp,rn) \qquad (7)$$

$$\kappa_I = |C| / gm_2(rp,rn) = [Rec\text{-}Bias]/[1\text{-}Prev] = \Delta R \qquad (8)$$

Note however that Hand's approach doesn't make use of a chance-correct measure, and there is no longer a clear diagonal representing chance, or isocost lines representing equal cost. AUK [16] was introduced to average a chance-correct Kappa measure over fpr, specifically rejecting the distributional assumptions of Hand [17]. Similarly proponents of ROC [18] reject the distribution and assumptions of Hand as unnatural and show the consistency of ROC AUC under slightly different assumptions that do not impose a guessed cost distribution like Hand's H-measure/H-AUC [17].

Since Hand's approach makes complex distributional assumptions we find arbitrary, and is not so much intended to be plotted as used in calculating an AUC, we do not illustrate an H-curve here. As discussed earlier, the AUC is of limited value for classifier evaluation as it mixes up consistency of a classifier with its informedness.

### *LIFT Chart*

Another kind of chart similar to PN (and ROC) is LIFT, which is essentially TP (or tpr if we normalize) plotted against a different variable on the x-axis, pp rather than FP (or fpr).

LIFT emerged independently in Data Mining as a way of depicting the returns on a mail out relative to the number mailed [19], and is also illustrated in Fig. 1. It is actually very sensible, and worthy of broader use, in that it effectively plots recall against the number of positive labels accepted. This situation is very similar to the Information Retrieval decision about how many search results to display or view, although we can also determine it via a threshold. LIFT is thus theoretically attractive as it avoids any dependency on the outcomes of the system (which Hand objected to), and the count or proportion of positives predicted corresponds to a credibility threshold on the classifier (but it does represent a uniform distribution, which Hand also objected to, but LIFT uses the actual empirical distribution rather than the arbitrary prior Hand proposes).

In fact, many classifiers allow direct estimation of a probability distribution on predictions, and both ROC curves and LIFT charts can be used to shape any credibility or distance measure into an empirical probability distribution, as we can adjust a threshold or vary a parameter – and then just count the positives.

### *BIFT Chart*

Clearly we can also make the same transformation we did for ROC → BOC and plot Bookmaker informedness, tpr-fpr, instead of tpr, so we get LIFT → BIFT (see Fig. 2).

### *Independence of AUC*

Concerns have been noted above [17] about the averaging over a distribution over the errors, but the following two theorems demonstrate that there is no need for concern. It turns out that optimizations of AUC for ROC, BOC, LIFT and BIFT are all equivalent.

**Theorem 1A**. The probability of an informed decision for +ves (tpr-fpr) and –ves (tnr-fnr) are the same, and the areas under the ROC curve (tpr vs fpr) and the dual ROC' curve (tpr vs fpr) are the same, and the areas under the BOC curve (tpr-fpr vs fpr) and the dual BOC' curve (tnr-fnr vs fnr) are the same.

**Proof Insight**. In the dichotomous case, a single decision determines both whether you think it is positive and whether you think it is negative. Similarly tpr+fnr = tnr+fpr =1, so the dual of ROC graph is simply a reflection, so AUROC = AUROC'. The area under the chance line is ½, so when it is subtracted off we have AUBOC = AUBOC' = Gini/2.
**Formal Proof**. Result was discovered by simplification of the Bookmaker cost formula.

**Theorem 1B.** The area under the BIFT curve is also AUBOC = AUROC – ½ = Gini/2.

**Proof Insight**. Every new example included as we reduce the threshold adds a new segment to the BIFT chart, incrementing its area, but also increments either the BOC chart or its dual BOC' chart. Considering the normalization of the chart x-axes, we have the equation N*AUBIFT = RN*AUBOC + RP*AUBOC'. Result follows now from 1A.
**Formal Proof**. Result was discovered by straightforward simplification of the integral.

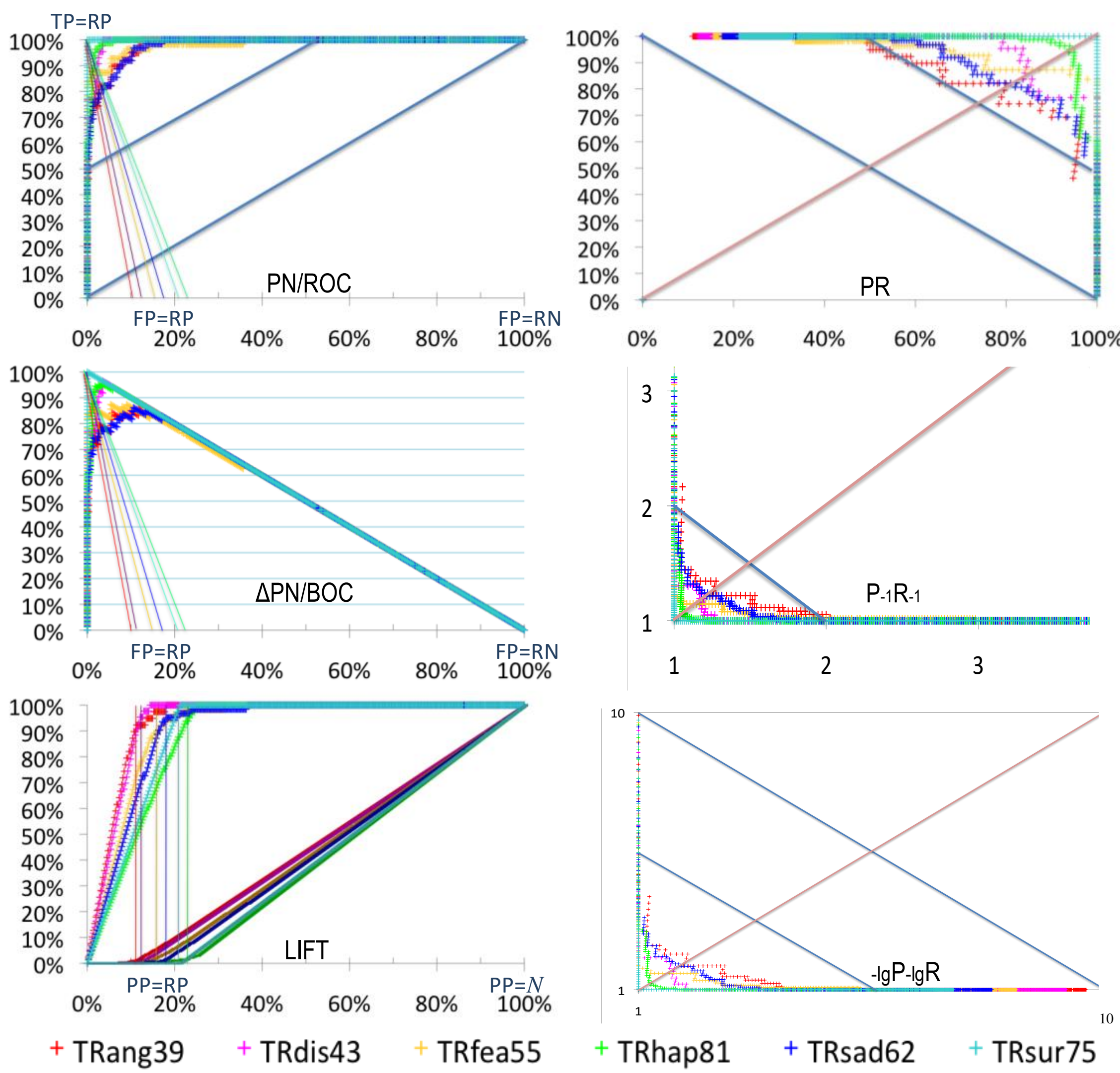

**Figure 1. Comparison of PN/ROC, ΔPN/BOC, PR, and LIFT – plotting Recall for each of 6 classes.**

**PN/Receiver Operating Characteristics (PN:TP vs FP; ROC:tpr vs fpr) and ΔPN /Bookmaker Operating Characteristics (ΔPN:TP-FP vs FP; BOC:tpr-fpr vs fpr) graphs show blue equal Informedness (tpr=fpr) isobars and thin class-color break-even lines. Areas under the curve, AUC for ROC and AUK=Gini for BOC, represent respectively the probability that a positive will be ranked higher than a negative, and half the probability that an informed decision will be made, averaged by threshold.**

**LIFT shows Recall against Bias so the class-color break-even lines are vertical (Bias=Prev) and the chance line (tpr=fpr=1-InvRecall) becomes the class-color curveset below the main diagonal while the area between the curves represents average Informedness or Gini components of Bookmaker.**

**Precision-Recall (PR) graphs Recall against Precision showing blue-gray Arithmetic Mean isobars and a red-brown break-even line. The reciprocal Precision-Recall ($P_{-1}R_{-1}$) has Harmonic Mean isobars (corresponding to F-measure), while information domain PR ($_{-lg}P_{-lg}R$) has Geometric Mean isobars. The shapes are similar as the scaling is near linear for small errors as $-\lg(1-\varepsilon) \simeq \lg(1+\varepsilon) \simeq \varepsilon$ for $\varepsilon \ll 1$.**

**Data is from Multi-Classifier Fusion for Facial Emotion Recognition with Cohn-Kanade dataset [11]. True [Positive] Rate results are shown for anger, disgust, fear, happiness, sadness and surprise based on 10-fold Cross Validation, with key showing the numbers of images per real class.**

## III. The Multiclass Case

### *Measures with K>2 Classes*

The measures defined so far have all been either intrinsically based on evaluation of a single class (Recall, Precision, F-Measure) or have been presented in the context of a dichotomous or two-class problem (Accuracy, AUC, Informedness, AUK, …).

We have noted that appropriately weighted averages of Recall and Inverse Recall, or Precision and Inverse Precision, give Accuracy. This extends to the multiclass case too. We can sum the Recall for the different classes weighted by the associated prevalences, or we can sum the Precision for the different classes weighted by the associated biases, giving Accuracy. Similarly we can sum Accuracy over multiple datasets weighted by the associated proportions (macro-averaging). Multiclass Accuracy is a macro-average of Recall (based on numbers of positive cases) or Precision (based on numbers of positive predictions), but F-measure is calculated as their harmonic mean.

Informedness relates to a particular prediction or 'bet' and our total Informedness or 'winnings' must be averaged over the proportion of the time we made the bet at each price, that is averaged over the bias of the system. This result is thus equivalent to the Bookmaker Informedness for the multiclass case, Markedness must similarly be averaged weighted by prevalence, while Correlation is calculated as their geometric mean [15].

The Cohen and Fleiss Kappa are similarly generalizations of a simple dichotomous case to a more general case of multiple classes, and shouldn't be macro-averaged [15].

### *Visualizations with K>2 Classes*

Figure 1 has illustrated the dichotomous measures and tradeoffs using as an example a 6-class problem in emotion recognition [20], smooth and detailed as a result of being the (cross-validated) fusion of multiple base classifiers. The Precision-Recall (PR) graph is noteworthy compared to the ROC, BOC and LIFT charts in that the break-even lines coincide for all 6 classes, when Precision=Recall. Different slope lines indicate different weightings of Precision vs Recall, or equivalently different Bias/Prevalence ratios.

Since the 6 distributions come from the same dataset, a choice of Bias>Prevalence for one class necessarily entails a setting of Bias<Prevalence for another, and the isoskew lines then diverge. The deviations from the skew implied by the prevalences are equivalent to the skews implied by different explicit cost choices, which in turn optimize with different bias skews. The particular case of interest here is the relative difference in the bias of two classifiers, or between a classifier bias and the corresponding class prevalence. This skew ratio is what we are looking at with these isoskew lines, and describes the relative drift (RD) of a classifier away from the true prevalence. The basis for our multiclass visualizations will be a variant of LIFT that plots Bookmaker Informedness directly for each individual class, BIFT, rather than displaying it implicitly as the difference between the Recall and Inverse Recall curves shown in Fig. 1. BIFT is illustrated for the same data in Fig. 2.

The LIFT and BIFT charts are arguably clearer than ROC or BOC as they spread out the convergent isoskew lines to be parallel, and thus spread out the graphed function around the optimum. This avoids the clutter and lack of clarity near the peak for BOC and ROC. PR spreads them out further to be divergent around the optimum, as it is specifically exploring the tradeoff in Bias versus Prevalence (in the denominators of Recall and Precision respectively).

Since there are other advantages to the ROC and LIFT type charts, and in particular we are interested in chance-correct assessment rather than biased measures (as highlighted specifically in the transformations ROC → BOC and LIFT → BIFT), we will explore additional transformations that *both* retain this key advantage of ROC and LIFT *and* address the two striking difference we observe when we compare to the PR chart:

a. PR unifies the Bias=Prevalence isoskew line; and

b. PR allows a high resolution view around the optimum.

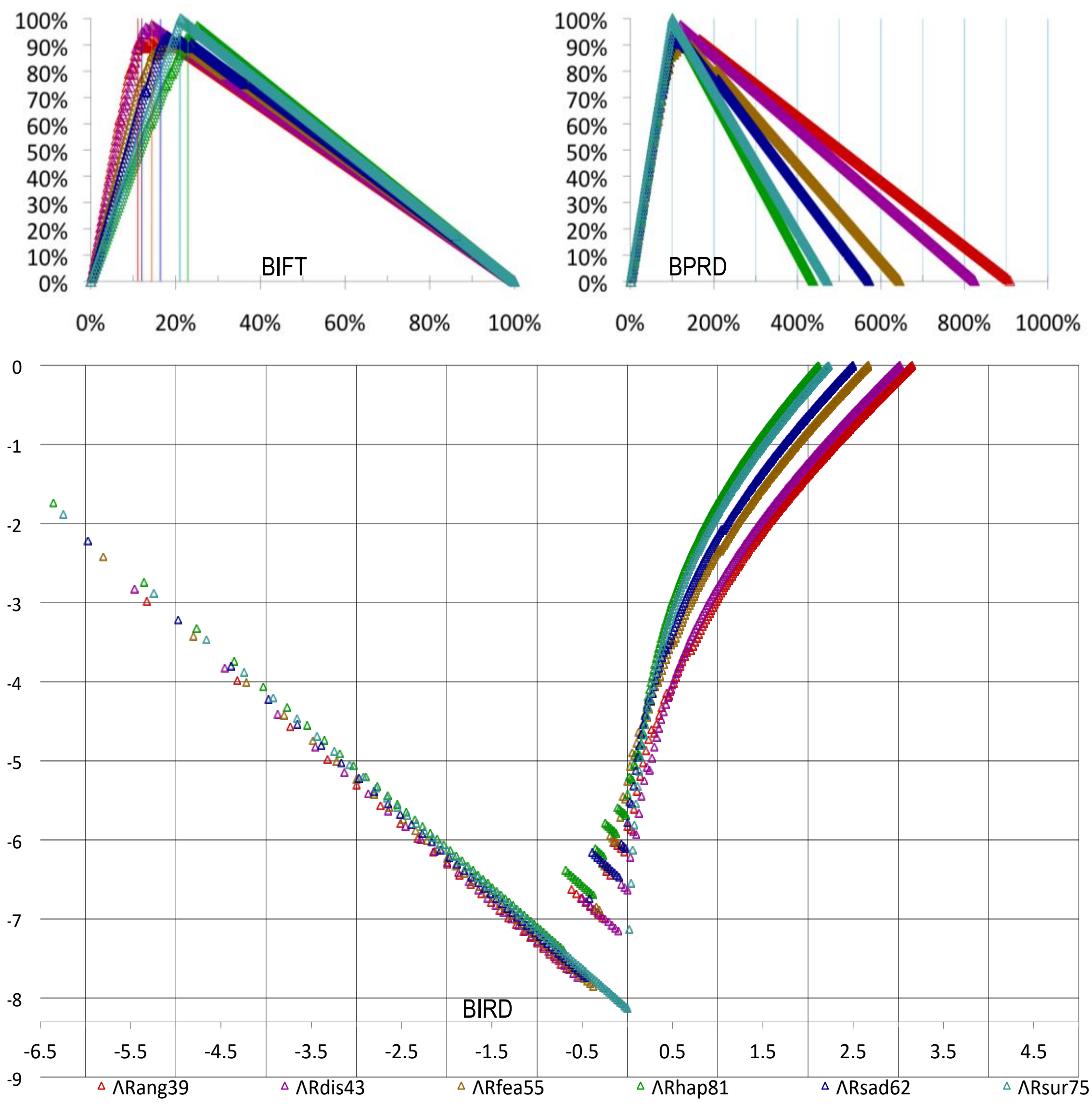

**Figure 2. Comparison of BIFT, BPRD and BIRD**

**BIFT Is a chance-corrected variant of LIFT based on Bookmaker Informedness, plotting tpr-fpr = Recall-Fallout = Recall + Inverse Recall – 1 rather than Recall and Fallout separately as in Fig. 1. Note drift to *right* of isoskew lines due to lack of Precision (more predictions needed to find all).**

**BPRD plots Bookmaker Probability = tpr-fpr of an informed decision versus the Relative Drift. Note drift to *right* of 100% isoskew meridian and blip to the right of 200% for two confused classes.**

**BIRD plots Bookmaker Information = H(stpr) – H(sfpr) versus the log Relative Drift.**
**Note optimum Bookmaker Probability or Information corresponds to a minimum Information Loss.**
**Note that smoothing means that stpr=1, sfpr=0 is not permitted to occur for infinite Information Loss.**
**Noise/error FNs "cut the corner" and shift any optimum *left* before break-even point at 0 (x-axis).**
**This BIRDhead emphasises selection of (wrong) members of the preponderant other classes.**

### *1) DoF – Pareto Optimization*

The problem with visualization of any evaluation measure for multiple classes, is that there are K-1 degrees of freedom to explore, amplifying the difficulty of our pareto optimization problem as K increases beyond 2. Basically, we have a variety of mechanisms to optimize an individual class or dichotomy, and pareto optimization replaces the expectation of finding an overall optimum directly with the idea of factoring out a dominated solution space that cannot possibly contain the optimum.

The thresholds of all but one class could be set independently; for example, leaving the last class to suffer whatever fate is left to it. This is in fact what has happened with the measures and visualizations based on Recall and Precision – how well the negatives are handled is only indirectly taken into account, if at all. It is clearly difficult to plot Accuracy or Informedness against the multiple parameters that can be varied, and setting the optimum thresholds becomes a significant optimization problem in its own right. Indeed we have a fusion problem which is a substantial problem for classifiers (like SVM) that need to combine multiple classifiers together to solve multiclass problems (either one vs rest for O(K) or pairwise for $O(K_2)$ classifiers to fuse). And it is precisely this problem we are addressing if we want to use visualization to help us second-guess the classifier, and go beyond simple voting and bias heuristics as we explore better combinations of parameters or different conditions (costs or prevalences).

### *2) BPRD – Bookmaker Probabilty vs Relative Drift*

The Bias = Prevalence isoskew line in the PR chart shows us one consistent system based on this heuristic, but the ROC and BOC charts don't due to their different slopes when the problem is unbalanced, that is the different classes have different prevalences as in the example used for Fig. 1. The LIFT chart is no better… but, being parallel, an appropriate rescaling can make these break-even isoskew lines coincide. The scaling we require simply compensates directly for the relative drift (RD=Bias/Prev=PP/RP).

Figure 2 shows the BIFT modification of LIFT, showing the Informedness ΔP'=tpr-fpr = Recall + InverseRecall – 1, so that the area under the curve corresponds to the area between the curves in the LIFT chart, analogous to the ROC to BOC display of ΔR for each class (where we avoid the inconvenient prime and indicate this derives from Recall). Figure 2 next shows the BPRD which is also based on Bookmaker Informedness but compensates for relative drift (RD). In this chart 100% corresponds to the number of positive predictions that should be made to match the number of positive classes, and we are plotting the gain according to Bookmaker odds versus the relative drift multiplier. Here Bookmaker is estimating Probability of an informed decision (BP). Thus

Recall=Precision corresponds to the vertical line given by RD=100%. But there is another currency or cost-basis we can consider, pricing on the basis of Shannon Information rather than traditional probabilistic Bookmaker odds.

### *3) BIRD – Bookmaker Information vs Relative Drift*

Close inspection of the PR graph will show that it is very noisy near an optimum, and the apparent optima for each class can be far from the break-even line. The optima we see in this graph relate to an arithmetic mean (A) of Recall and Precision, but taking the reciprocals of the axes ($P_{-1}R_{-1}$), so it optimizes the harmonic mean (F-measure), doesn't change this effect for the lower prevalence classes. What we are seeing here is fairly arbitrary noise effects as we increment, as we find each apparent true positive amongst a sea of false positives. What is more important is the size of this boundary region, where we cut the corner and miss the optimum that is achieved so sharply for the most prevalent class, and a graph that zooms in may be more useful than the reciprocal and log transforms implicit in the unbounded F- and G-measures.

The geometric mean (G-measure), is arguably most appropriate, but can't actually move the bumps in the curves. It is however what is indicated for our chance-correct Informedness and Markedness equivalents, giving us Correlation. This corresponds to linear isometrics on a log scale version of the PR graph, against which the arithmetic mean and the harmonic mean become complementary, with $\log(A)+\log(F) = 2\log(G)$.

The log scale versions of the graphs moreover have an information-theoretic basis. Small variations in probabilities near 1 have small effects ($H(x) = -\log(1-x) \simeq x$). But we are interested in minimizing cost or maximizing gain (rather than Recall or Precision) and the information conveyed by a prediction under Bookmaker odds, is $H_I = -\log(\text{tpr}–\text{fpr})$. However there are problems with this, given Informedness can be 0 or even negative (if information is available but used incorrectly). We therefore cost using the information loss due to good predictability (tpr) versus poor predictability (fpr), giving an analogous information theoretic form to Bookmaker, just setting the prices differently. This should also be 0 for chance (tpr=fpr), that is if the information is not exploited, and is negative if information is utilized, positive if information is misused: $B_I = H(\text{tpr}) – H(\text{fpr}) = -\log(LR+)$, where LR+ is the Likelihood Ratio, tpr/fpr.

There is however a problem here – either or both tpr and/or fpr can be 0, and at the origin in the ROC curve they are both 0. This is also technically a problem with F-measure, as when tpr is 0, it means Recall and Precision are 0 and the harmonic mean is undefined (although by convention we define it as 0). This case also represents a technical hole in the claim that if one system dominates on the PR curve it will also dominate on the ROC curve and vice-versa [21].

We can exclude these cases from the curves, but more generally it is convenient to use Laplace (e.g. AddOne or AddHalf) smoothing to ensure that tpr and fpr stay non-zero (which is theoretically motivated by our finite sample size – we could at any point be just about to see another apparent true or false positive). This has the consequence of defining a finite best case for a smoothed fpr based on Laplace smoothing term s which is added to our counts (smoothed TP is denoted ${}_{s+}TP$=s+TP, etc.) and we can generalize to a smoothed Relative Density similarly. Thus we introduce a smoothed definition of Bookmaker Information:

$$stpr = {}_{s+}TP/{}_{s+}RP \geq 1/{}_{s+}RP \geq 1/N \text{ for } s=1 \qquad (9)$$

$$sfpr = {}_{s+}FP/{}_{s+}RN \geq 1/{}_{s+}RN \geq 1/N \text{ for } s=1 \qquad (10)$$

$$B_I = H(stpr) - H(sfpr) = -\log(stpr/sfpr) \qquad (11)$$

$$sRD = {}_{s+}PP/{}_{s+}RP \qquad (12)$$

$B_I = -\log_2(stpr/sfpr)$ versus $\log_2(sRD)$ thus gives us our final graph for Fig. 2, the appropriately shaped BIRD graph (Bookmaker Information vs Relative Drift). This information-theoretic graph shows us the significant variation from the desired optimum while hiding much of the noise visible in the graphs of Fig. 1. The limiting bias (reached for fpr=0) defined by our smoothing assumption is thus $-\log({}_{s+}RN)$, and spreads out the left wing with 'feathers' when prevalences are not balanced. Given enough data to mitigate the effect of smoothing, and a balanced set of prevalences across the K classes, the maximum relative drift is K and so $-\log_2(K)$ is an expected upper limit to the x-axis. In general, this upper bound reflects the skew as $-\log_2(RN/RP)$ when we predict exactly the negatives as positives. The more extreme cases will thus be seen for the lower prevalence classes, and so we see different coloured feathers at the right wing tip.

The BIRD head spreads out the errors: without the smoothing, the unsmoothed FP=0 case (left wing) saves 'infinite' information but smoothing takes into account that our error estimate is based on a finite sample (0 errors only means infinite information if we believe that this means an error is impossible and thus would convey infinite information). With the smoothing, FP>0 (right wing) to FP=0 (left wing) still represents a significant jump, that is still enough that the optimum will be *left* of the balance meridian for Bookmaker Information costing (need to underpredict), rather than *right* with the Bookmaker Probability costing (so want to overpredict, but can't for *all* classes). Generally, if small specific 'positive' classes are overpredicted (general populace isn't).

# IV. Conclusions

In the PR and ROC graphs, we have a single point representing the best achievable point for any classifier, but ROC is interpretable independently of the prevalences of the individual classes, while PR operating points reflect bias and prevalence – the higher the skew towards positives, the easier it is to achieve high precision, while high bias towards positive predictions will automatically achieve high recall, pushing operating points towards the (1,1) point irrespective of the performance of the underlying classifier. For example, always guessing positive achieves 100% Recall, and Precision → 100% automatically as Prevalence → 100% - even though we are just guessing. With ROC guessing tends to turn up the same proportion of correct labels (tpr) as incorrect labels (fpr) so tpr = fpr is the chance line, and BOC or AUK makes this the axis so guessing scores 0.

Rotating the ROC graph into a BOC graph – making the diagonal the axis, has the additional advantage of expanding out the best achievable point to an entire axis – the top line of the graph. Thus instead of competitive solutions all bunching together in the corners as with ROC and PR, they are spread out, allowing better discrimination and tuning around the optimum. However, the x-axis still represents an error rate, which is a major criticism of ROC and its derivatives [17]. LIFT and BIFT replace the error rate by the positive prediction rate, or bias – something that we can control without supervision.

In the BIRD graph we now have the desirable property of parallel vertical isoskew lines, and, moving left or right of the central y-axis in Fig. 1, successive meridians indicate the loss of an additional bit of information relative to the optimal bias implied by prevalence.

If any class occurs more than it should (shifted right), the others must on average occur less (shifting left), according to the Bias = Prevalence model. If we do not want to assign equal cost to each class (in aggregate) as ROC and BIRD do by default, this would again shift and split the isoskew lines, and it would be appropriate to normalize the relative drift scale taking into account both cost and prevalence instead.

If the 'gold standard' is poor, the prevalences apparent in the 'real' class labels (and the identified false positives and negatives) can be wrong. If one class is more difficult to identify than another, if it has been undertrained or is subject to measurement error or label noise, the Bias = Prevalence model will be suboptimal. But these problems can be detected from the BIRD graph. The range of RD shown represents an order of magnitude discrepancy in Bias vs Prevalence.

Overall, we find the BIRD visualization most appropriate for selecting not just separate ‘optima’ for independent curves but a set of consistent predictions for each class (the counts must sum to *N*). We maximize classification of pathological cases versus the normal or default class both maximizing Informedness for those classes and minimizing information loss relative to the null hypothesis of normality. Moreover, the log scales emphasize how many bits of information loss we are penalized as we move away from the settings implied by the prevalence and/or cost structure of the dataset/application.

The BPRD and BIRD plots based on relative density are more convenient than ROC, LIFT and BIFT charts where percentages along the x-axis are only interpretable with reference to the actual known prevalences.

While the probability-scale BPRD normalization shows differences relative to prevalence, and unifies the left sides of the subtend triangular curves, it leads to spreading out of the range of overprediction in the right portion of the curves according to the prevalence of the *other* classes. With BIRD we see extension without this systematic spread, allowing significant (pareto) differences to be seen clearly, and with prevalence variation conveniently spreading out parallel layers in the left and central parts, and showing up as only linear extension on the right side. In our example we see consistent inflection at the 0 meridian, changing from linear (FP=0) to concave down (FP>0). Note that the 1st and 2nd derivatives of -log(p) are $-p^{-1}$ and $p^{-2}$, with p= +1 in the linear part of the left wing (high probability of correct prediction) and p≃ –1 in the right extrema (high probability of incorrect prediction). In relation to the centre part of the curve, the BIRD head, this can be expected to be dominated by normally distributed (or more accurately multinomial) noise or error that effectively selects between the positive (correct) and negative (incorrect) distribution – but which don’t simply sum.

We finally reiterate that averaging AUC across multiple one vs rest ROC curves to give a composite AUC or VUS result, does not provide a visual solution, it takes a single number derived from multiple binary graphs and combines blindly into a single number. This is only interpretable as a probability of weighting the correct class against others if the class-AUCs are macro-averaged with *bias* not *prevalence* weighting, as interpretation of the macro-average of the operating points’ dichtomous Informedness (=2AUC–1), as multiclass Informedness requires using *bias* weighting. This is because we have no oracle to identify which class we have and hence which classifier to use, rather we use the predicted class to make our decisions and thus must treat that as the positive class. This *bias*-weighted average was shown to be *optimal* [22], and other weighting systems were shown to be *suboptimal*. Similarly, caution is emphasized regarding the misleading use of AUC measures rather than Informedness [10].

## Sidebar 1. How to draw the graphs

**Precision and Recall and the PR graph are well known, we simply plot Recall against Precision based on Equation 1. However, the other graphs are less well known and we introduce some new variants. So we summarize the details here from the perspective of how to draw them. It is relatively easy to draw the plots using Matlab, and slightly more tricky to draw them in Excel, which is how Fig. 1 and Fig. 2 were generated. So we explain this for Excel, and offer downloadable spreadsheets to illustrate. It is assumed that the spreadsheet has, for each example y, its true class x, and a prediction value $f_{yz}$ for each possible label z, that is used to determine which value of z is predicted.**

**The tricky part of Excel is the limited number of standard functions available, e.g. for taking ranks or finding argmax or calculating counts under various conditions. To do this in a uniform way that works across all versions, we exploit the array formula capability – this allows formulae that work on whole arrays of cells, including performing arithmetic operations on cells (which can be stored in an appropriately shaped region) or summing over an array (to produce a result for a single cell). To indicate it is an array formula it is necessary to use Control-Shift-Enter rather than just Enter when entering a formula. We use --(Bool) to turn a Boolean relation into a 0-1 result which we can count by adding (|$f_{yz}$ >θ|), using multiply to achieve a 'logical and' with a second condition (|$f_{yz}$>θ & z=x|).**

**We use this technique to count the number of items about a threshold θx ($f_{yz}$ |>θx|) set for each output from classifier x (giving Px corresponding to PP for label x at each θx). If subject to the desired constraint about being true (to give TRx) or false (to give FRx) we have effectively TPR and FPR for the class x. We now show the formulae used to define the values for each axis and curve for each member of the family of ROC and LIFT graphs we presented (in probability form). As we have six classes, we don't talk about positives versus negatives as such, but tpr becomes TRx, fpr is FRx, pp is Px, and rp is Rx, for each class label x. Note that we use upper case here so that lower case can represent the label legibly without the use of tiny subscripts in the legend.**

**ΣPx represents the bias *skew* versus prevalence, and ΛPx is its base 2 *logarithm* (information *loss*) while THx represents the *entropy* associated with TRx (Recall for class x). H(p) = $-\log_2 p$ is the information or entropy associated with an event of probability p, with THx = H(sTRx), FHx = H(sFRx). The smoothed FRx, sFRx = $_{s+}$|>θx & –ve| / $_{s+}$|–ve|, uses the count $_{s+}$|Bool| = s+|Bool| for Laplace smoothing (we use s=1 for AddOne smoothing of the cardinality of the set of instances satisfying the Boolean condition/function).**

**ΔRx is the expected Bookmaker gain under the default ROC assumption of cost as inverse prevalence, while ΔHx is expected Bookmaker gain under an information theoretic cost model (expressed negatively corresponding to a net information loss for the predicted event).**

<table>
<tr><th>Chart</th><th>x-axis</th><th>y-axis (curves)</th></tr>
<tr><td>ROC</td><td>FRx = |>θx & –ve| / | |–ve|</td><td>TRx = |>θx & +ve| / |+ve|</td></tr>
<tr><td>BOC</td><td>FRx = |>θx & –ve| / | |–ve|</td><td>ΔRx = TRx – FRx</td></tr>
<tr><td>LIFT</td><td>PPx = |>θx| / <i>N</i></td><td>TRx = |>θx & +ve| / |+ve|</td></tr>
<tr><td>BIFT</td><td>PPx = |>θx| / <i>N</i></td><td>ΔRx = TRx – FRx</td></tr>
<tr><td>BPRD</td><td>ΣPx = $_{s+}$|>θx| / $_{s+}$|+ve|</td><td>ΔRx = TRx – FRx</td></tr>
<tr><td>BIRD</td><td>ΛPx = $\log_2$ΣPx</td><td>ΔHx = THx – FHx</td></tr>
</table>

## References

[1] C. Drummond, W. Elazmeh, N. Japkowicz, and P. Cochairs, (2006) *AAAI Workshop Evaluation Methods for Machine Learning*, WS-06-06, AAAI press, 2006.

[2] W. Klement, C. Drummond, N. Japkowicz, and S. Macskassy, (2009) *The Fourth Workshop Evaluation Methods for Machine Learning,* Proc. 26th Ann. Int'l Conf. Machine Learning (ICML '09), http://www.site.uottawa.ca/ICML09WS/index.html, 2009.

[3] R.C.Prati, G. Batista, M.C. Monard (2011) A Survey on Graphical Methods for Classification Predictive Performance Evaluation, *IEEE Trans Know Data Eng*. 23:11

[4] D.J. Hand (2001). A simple generalisation of the area under the ROC cruve for multiple class classification problems. *Machine Learning* 45:171-186.

[5] F. Provost and P. Domingos, Well Trained PETs: Improving Probability Estimation Trees, CeDER Working Paper IS-00-04, Stern School of Business, New York Univ., 2001.

[6] T. Landgrebe and R. Duin (2006). "A simplified extension of the area under the ROC to the multiclass domain" 17th annual symposium of the pattern recognition association of South Africa.

[7] C. Ferri, J. Hernandez-Orallo, and M. Salido, "Volume Under the ROC Surface for Multi-Class Problems," Proc. 14th European Conf. Machine Learning, pp. 108-120, 2003.

[8] D. Edwards, C. Metz, and R. Nishikawa, "The Hypervolume under the ROC Hypersurface of 'Near-Guessing' and 'Near- Perfect' Observers in N-Class Classification Tasks," IEEE Trans. Medical Imaging, vol. 24, no. 3, pp. 293-299, Mar. 2005.

[9] A. Bradley, "The Use of the Area under the ROC Curve in the Evaluation of Machine Learning Algorithms," Pattern Recognition, vol. 30, no. 7, pp. 1145-1159, 1997.

[10] D.M.W. Powers (2012), ROC-ConCert: ROC-Based Measurement of Consistency and Certainty, (2012) *Spring Congress on Engineering and Technology (SCET)* **2**:238-241 IEEE

[11] R. Baeza-Yates, B. Ribeiro-Neto (1999), *Modern Information Retrieval*, Addison Wesley.

[12] B. Schölkopf, R.C. Williamson, A.J. Smola, J. Shawe-Taylor, John C Platt (1999), Support Vector Method for Novelty Detection. *Neural Information Processing Systems (NIPS)* **12**:582-588

[13] P.A. Flach (2003). The Geometry of ROC Space: Understanding Machine Learning Metrics through ROC Isometrics, *International Conference on Machine Learning*, pp.226-233.

[14] P. Perruchet, R. Peereman (2004). The exploitation of distributional information in syllable processing. *Journal of Neurolinguistics* **17**:97-119.

[15] D.M.W. Powers (2012), The problem with Kappa, *Conference of the European Chapter of the Association for Computational Linguistics (EACL)*, pp.345-355

[16] U. Kaymak, A. Ben-David, R. Potharst (2012), The AUK: A simple alternative to the AUC, *Engineering Applications of Artificial Intelligence* **25**(5):1082–1089

[17] D.J. Hand (2009), Measuring classifier performance: a coherent alternative to the area under the ROC curve, *Machine Learning* **77**:103–123.

[18] P.A. Flach, J. Hernandez-Orallo and C. Ferri (2011), A Coherent Interpretation of AUC as a Measure of Aggregated Classification Performance, *International Conference on Machine Learning* (ICML), 139-146.

[19] M.J.A. Berry and G. Linoff (1997), *Data mining techniques for marketing, sales, and customer support*, New York: Wiley.

[20] X.B. Jia, Y.H. Zhang, D.M.W. Powers and H.B. Ali (2014), Multi-classifier fusion based facial expression recognition approach, *KSII Transactions on Internet & Information Systems* **8**(1):196-212.

[21] J. Davis; M. Goadric (2006). "The Relationship Between Precision-Recall and ROC Curves", *ICML*, pp.233-240. Article (CrossRef Link).

[22] D.M.W. Powers (2003), Recall & Precision versus The Bookmaker, *International Conference on Cognitive Science* (ICCS), 529-534.